\documentclass[final,3p,times,twocolumn,authoryear]{elsarticle}

\usepackage{times}
\usepackage{graphicx}
\usepackage{latexsym}

\usepackage{color}
\usepackage{amsmath}
\usepackage{booktabs}
\usepackage{algorithm}
\usepackage{algorithmic}
\usepackage{subfigure}
\usepackage{multirow}
\usepackage{amssymb}
\usepackage{bm}
\usepackage{color}
\usepackage{booktabs}
\usepackage[colorlinks,
            linkcolor=blue, 
            anchorcolor=blue,  
            citecolor=blue,     
            ]{hyperref}

\def\a{\bm{a}} 
\def\b{\bm{b}} 
\def\C{\bm{C}} 
\def\c{\bm{c}} 
\def\E{\bm{E}} 
\def\e{\bm{e}} 
\def\h{\bm{h}} 
\def\o{\bm{o}} 
\def\l{\bm{l}} 
\def\m{\bm{m}} 
\def\n{\bm{n}} 
\def\p{\bm{p}} 
\def\g{\bm{g}} 
\def\u{\bm{u}} 
\def\W{\bm{W}} 
\def\x{\bm{x}} 
\def\y{\bm{y}} 
\def\model{Bi-GTPPP } 
\def\setU{\mathbb{U}} 
\def\setC{\mathbb{C}}
\def\setR{\mathbb{R}}

\journal{Neural Networks}

\begin{document}

\begin{frontmatter}



\title{Exploiting Bi-directional Global Transition Patterns and Personal Preferences for Missing POI Category Identification}


\author[a,c]{Dongbo Xi}
\ead{xidongbo@meituan.com}
\author[b,a]{Fuzhen Zhuang\corref{cor1}}
\ead{zhuangfuzhen@ict.ac.cn}
\author[e]{Yanchi Liu}
\author[f]{Hengshu Zhu}
\author[g]{Pengpeng Zhao}
\author[h]{Chang Tan}
\author[a,d]{Qing He}

\cortext[cor1]{Corresponding author: Fuzhen Zhuang}

\address[a]{Key Lab of Intelligent Information Processing of Chinese Academy of Sciences (CAS),\\ Institute of Computing Technology, CAS, Beijing 100190, China
}
\address[b]{Xiamen Data Intelligence Academy of ICT, CAS, China}

\address[c]{Meituan-Dianping Group, China}
\address[d]{University of Chinese Academy of Sciences, Beijing 100049, China}
\address[e]{Management Science \& Information Systems, Rutgers University, USA}
\address[f]{Baidu Inc., Beijing, China}
\address[g]{Soochow University, China}
\address[h]{iFLYTEK, China}

\begin{abstract}
Recent years have witnessed the increasing popularity of Location-based Social Network (LBSN) services, which provides unparalleled opportunities to build personalized Point-of-Interest (POI) recommender systems. 
 Existing POI recommendation and location prediction tasks utilize past information for future recommendation or prediction from a single direction perspective, while the missing POI category identification task needs to utilize the check-in information both before and after the missing category.
 Therefore, a long-standing challenge is how to effectively identify the missing POI categories  at any time in the real-world check-in data of mobile users. To this end, in this paper, we propose a novel neural network approach to identify the missing POI categories by integrating both bi-directional global non-personal transition patterns and personal preferences of users. Specifically, we delicately design an attention matching cell to model how well the check-in category information matches their non-personal transition patterns and personal preferences. Finally, we evaluate our model on two real-world datasets, which clearly validate its effectiveness compared with the state-of-the-art baselines.
Furthermore, our model can be naturally extended to address next POI category recommendation and prediction tasks with competitive performance.
\end{abstract}




\end{frontmatter}



\section{Introduction}
The rapid development and increasing popularity of Location-based Social Network (LBSN) services encourage more and more users to share their real-life experiences. Data collected by various LBSN services have been effectively leveraged for studying users' online activities and mobility patterns, which provide unparalleled opportunities to built personalized  POI recommender systems.
Generally, the user's historical visiting records can be regarded as a set of check-ins that each one contains a POI, a timestamp, a POI category, and so on. Indeed, the category information is conducive to explaining users' activities and plays a crucial role in various POI-oriented tasks.

Most of the existing POI-oriented studies mainly focus on recommending or predicting the category of POI where a user may go in the future or improve the performance of next POI recommendation.
Recently, considerable efforts have been made to address next POI category prediction \citep{zhang2017large}, and utilize the category information to solve POI recommendation and location prediction problems \citep{chen2015information,he2017category,ye2013s}.
However, these methods utilize past information for future prediction or recommendation from a single direction perspective, while the missing POI category identification task needs to utilize the check-in information before and after the missing category, which naturally calls for a bi-directional solution.
Therefore, a long-standing challenge is how to effectively identify the missing POI categories  at any time in the real-world  check-in data of mobile users.


Along this line, we propose a novel \textbf{Bi}-directional \textbf{G}lobal \textbf{T}ransition \textbf{P}atterns and \textbf{P}ersonal \textbf{P}references model named \model for missing POI category identification. The proposed model takes both non-personal and personal preferences into consideration. On the one hand, users' check-in activities usually have some public transition patterns, which are non-personal preferences. For example, users often go to dinner after work and watch a movie after dinner. The transition patterns $work\rightarrow dinner, dinner\rightarrow movie$ are global and non-personal for all users. Our  model is designed to capture bi-directional global transition patterns. On the other hand, different users have personal preferences that affect the check-in behaviors of users. 
These two preferences are integrated through a delicately designed attention matching cell between the output of LSTM network \citep{hochreiter1997long} and the preferences. 
With the bi-directional global transition patterns and users' personal preferences, the \model can yield more accurate missing POI category identification.

The main contributions of this work are listed as follows:
\begin{itemize} 
\item  The proposed model can address the non-trivial missing POI category identification task via utilizing bi-directional global transition patterns and users' personal preferences, and existing next POI category prediction methods are not suitable for our task.
\item The existing next POI category prediction task can be seen as a special case of the missing POI category identification task. The proposed model can be easily extended to address the next POI category prediction task by only using forward sequence information, while the existing prediction models can not address the identification task well.
\item Based on the experiments conducted on real-world datasets, the proposed \model model achieves significantly better performance compared with existing various state-of-the-art baselines.

\end{itemize}

\section{Related Work}
In POI-oriented studies, two important tasks are POI recommendation and location prediction.
In this section, we present the related work in threefold: general POI recommendation and location prediction, category-aware POI recommendation and location prediction, and neural network-based methods.

\subsection{General POI Recommendation and Location Prediction}
Factoring Personalized Markov Chains and Localized Regions \citep{cheng2013you} takes users' movement constraint into account via exploiting the personalized Markov chain in the check-in sequence. Personalized Ranking Metric Embedding (PRME) \citep{feng2015personalized} integrates sequential information, individual preference, and geographical influence to improve the recommendation performance. Graph-based Embedding \citep{xie2016learning} jointly captures the sequential effect, geographical influence, temporal cyclic effect and semantic effect by embedding into low dimensional space. More informations such as temporal effects \citep{gao2013exploring}, spatial-aware \citep{yin2017spatial}, behavior patterns \citep{he2016inferring}, various contexts \citep{yang2017bridging} also have been studied accordingly in POI recommendation and location prediction tasks.

\subsection{Category-aware POI Recommendation and Location Prediction}
Real-life POI-oriented tasks usually suffer from huge search space, which is because the number of POIs is large and needs a lot of computational costs, 
while the POI category can help filter candidate POIs and thus reduce the search space for efficiency and improve the recommendation performance.
Context-Aware POI Category Prediction \citep{zhang2017large} emphasizes the significance of category information in large-scale POI recommendation. More and more efforts have been made to utilize the category information. Ye et al. \citep{ye2013s} proposed a framework which exploits region categories to predict the most likely location of users given their previous activities. Liu et al. \citep{liu2013personalized} employed matrix factorization to predict a user's preference on locations in the corresponding categories. A new POI recommendation problem, namely top-K location category based POI recommendation \citep{chen2015information}, has been formulated considering that users are more interested in tasting a wide range of location categories. Listwise Bayesian
Personalized Ranking approach \citep{he2017category} has been proposed to predict the category ranking to filter candidate POIs. Category information has also been considered in \citep{zhou2014probabilistic} and extended to more applications \citep{xiao2010finding,rodrigues2012automatic}.

\subsection{Neural Network for POI Recommendation and Location Prediction}
Neural networks have been used in the field of POI recommendation and location prediction \citep{liu2016predicting,wang2017deep,yin2017spatial}. For examples, a method called Spatial Temporal Recurrent Neural Networks (STRNN) \citep{liu2016predicting} was proposed to model temporal and spatial contexts in each layer. Heterogeneous features and spatial-aware personal preferences were utilized by Spatial-Aware Hierarchical Collaborative Deep Learning model \citep{yin2017spatial}. Some LSTM-based approaches \citep{zhu2017next,zhao2018go} try to capture short-term and long-term characteristics via specifically designed gates. User preference over POIs and context associated with users and POIs were predicted simultaneously in PACE (Preference And Context Embedding) \citep{yang2017bridging}. More neural network-based approaches \citep{wang2017deep,zhang2017next,yang2017neural} have also been adopted to address POI recommendation and location prediction.
However, the above methods are not specially designed for missing POI category identification and these efforts fail to capture both bi-directional global transition patterns and users' personalized preferences for missing POI category identification. So they can not address the identification task well. 

Besides, the most relevant task to missing POI category identification is the missing POI check-in identification.
The work \citep{xi2019modelling} is the first to address the missing POI check-in identification by modelling of bi-directional spatio-temporal dependence and users’ dynamic preferences (Bi-STDDP).
However, the missing POI category check-ins identification is different from the missing POI check-in identification in many ways.
Therefore, the model is not suitable for missing POI category check-ins identification task.

\section{Methodology}
In this section, we first formulate the problem of missing POI category identification, and then present the details of the proposed \model model, which contains three parts of attention matching cell, bi-directional global transition patterns, and personal preferences.

\subsection{Problem Statement}
Let $\setU=\{u_1,u_2,...,u_N\}$ is a set of $N$ users and $\setC=\{c_1,c_2,...,c_M\}$ is a set of $M$ POI categories. Each sample is associated with a category check-ins list of user $u$ $\C^u=\{c^u_{1},c^u_{2},...,c^u_{L}\}$, where $c^u_{l}$ means user $u$'s $l$-th check-in category is $c^u_{l}$. Assume the $l$-th check-in category $c^u_{l}$ of user $u$ is missing, the task is to identify which POI category the user $u$ visited according to the forward sequence $\{c^u_{1},c^u_{2},...,c^u_{l-1}\}$ and the backward sequence $\{c^u_{l+1},c^u_{l+2},...,c^u_{L}\}$. 

\subsection{Attention Matching cell}
Firstly, we formalize a delicately designed attention matching cell which is adopted in our model to weight different preference features:
\begin{eqnarray}
cell(\a,\b)&=&(1-s)\times\a+s\times\b\label{cell},\\
s&=&0.5+0.5\times\cos(\a,\b)\\
&=&0.5+0.5\times\frac{\a^\top\b}{\Vert\a\Vert\Vert\b\Vert},
\end{eqnarray}
where $\a$ is the feature extracted based on the existing data (e.g., check-in categories sequence) by LSTM network, $\b$ expresses the preferences which are global transition patterns or personal preferences in our model, and $\a$ and $\b$ are in the same space $\setR^M$. The weight $s$ is the normalized cosine similarity between $\a$ and $\b$,
and indicates how much $\a$ matches $\b$. If feature $\a$ matches preference $\b$ well, then preference $\b$ should have a bigger weight $s$, otherwise the feature $\a$ should be retained more. Note that $cell(\a,\b)$ is not equal to $cell(\b,\a)$.
 
\subsection{Bi-directional Global Transition Patterns}
In this subsection, we first extract features from users' check-in categories sequence, and then integrate global transition patterns.

Firstly, we capture POI category information with embedding layer. The embedding layer can be seen as performing the latent factor modeling for category popularity. It learns one matrix $\E_c$, each row of which represents a POI category. If we use one-hot encoded category $\c^u_{l-k}$, $\c^u_{l+k}\in\setR^M$ as input vectors, the outputs of embedding layer can be expressed as
\begin{eqnarray}
\e(\c^u_{l-k})&=&\E_c ^\top\c^u_{l-k},\\
\e(\c^u_{l+k})&=&\E_c^\top\c^u_{l+k},
\end{eqnarray}
where $1\leq k\leq w$, and $w$ is the window width, $\E_c\in\setR^{M\times d}$ denotes the embedding matrix for categories, $d$ is the dimension of embedding vectors.

LSTM \citep{hochreiter1997long} is capable of learning short and long-term dependencies and has become an effective and scalable model for sequential prediction problems, we use the basic LSTM to capture user's forward and backward check-in information:
\begin{eqnarray}
\l^u_{l-1}&=&LSTM(\e(\c^u_{l-w:l-1})),\\
\l^u_{l+1}&=&LSTM(\e(\c^u_{l+w:l+1})),
\end{eqnarray}
where $\e(\c^u_{l-w:l-1})$ and $\e(\c^u_{l+w:l+1})$ are the embedded forward and backward check-in sequences respectively, $\l^u_{l-1}\in\setR^h$ and $\l^u_{l+1}\in\setR^h$ are the bi-directional features extracted from users' check-in categories sequence by LSTM and $h$ is the dimension of the LSTM output space.

Next, we add a hidden layer to transform the output of LSTM network to another space for applying attention matching cell:
\begin{eqnarray}
\h^u_{l-1}&=&f(\W_{f}\l^u_{l-1}),\\
\h^u_{l+1}&=&f(\W_{b}\l^u_{l+1}),
\end{eqnarray}
where $\W_{f}\in\setR^{M\times h}$ and $\W_{b}\in\setR^{M\times h}$ are the parameters of transformation matrices.

Then, user's check-in activities usually have some public transition patterns, for example, users often go to dinner after work and watch a movie after dinner. The transition patterns $work\rightarrow dinner, dinner\rightarrow movie$ are global for all users. Our \model model is designed to capture the global transition patterns from bi-direction. For the missing category $c^u_l$, we need to capture the most relevant forward global transition patterns $\c^u_{l-1}\rightarrow\c^u_l$ and backward $\c^u_{l+1}\rightarrow\c^u_l$:
\begin{eqnarray}
\g_{l-1}&=&f(\E_f^\top\c^u_{l-1})\label{gf},\\
\g_{l+1}&=&f(\E_b^\top\c^u_{l+1})\label{gb},
\end{eqnarray}
where $\E_f\in\setR^{M\times M}$ and $\E_b\in\setR^{M\times M}$ denote the forward transition embedding matrix and backward transition embedding matrix respectively, 
the activation function $f(x)$ is chosen as a $\tanh$ function $f(x)=\frac{e^{x}-e^{-x}}{e^{x}+e^{-x}}$. 
The output $\g_{l-1}\in\setR^M$ can be normalized and expresses the transition distribution from category $\c^u_{l-1}$ to all candidate categories, and $\g_{l+1}\in\setR^M$ is the same. However, we don't do the normalization considering the expressive power.

Now, we can apply the delicately designed attention matching cell in Equation (\ref{cell}) to model how well the user's bi-directional check-in information matches the global transition patterns: 
\begin{eqnarray}
\m^u_{l-1}&=&cell(\h^u_{l-1},\g_{l-1}),\\
\m^u_{l+1}&=&cell(\h^u_{l+1},\g_{l+1}),\\
\m^u_l&=&\m^u_{l-1}+\m^u_{l+1}.\label{global cell}
\end{eqnarray}
The attention matching $cell$ interpolates $\h^u_{l-1}$ and $\g_{l-1}$, $\h^u_{l+1}$ and $\g_{l+1}$ respectively, and indicates how much $\h^u_{l-1}$ matches $\g_{l-1}$ or $\h^u_{l+1}$ matches $\g_{l+1}$. If user $u$'s forward check-in information $\h^u_{l-1}$ matches forward preference $\g_{l-1}$ well, then forward preference $\g_{l-1}$ should have a bigger weight, otherwise the feature $\h^u_{l-1}$ should be retained more, and $cell(\h^u_{l+1},\g_{l+1})$ is also the same. Finally, the bi-directional informations are added to capture the bi-directional global transition patterns $\m^u_{l}$.

\subsection{Personal Preferences and the Final \model Model}
\begin{figure*}[!t]
	\begin{center}
		\includegraphics[width=0.7\linewidth]{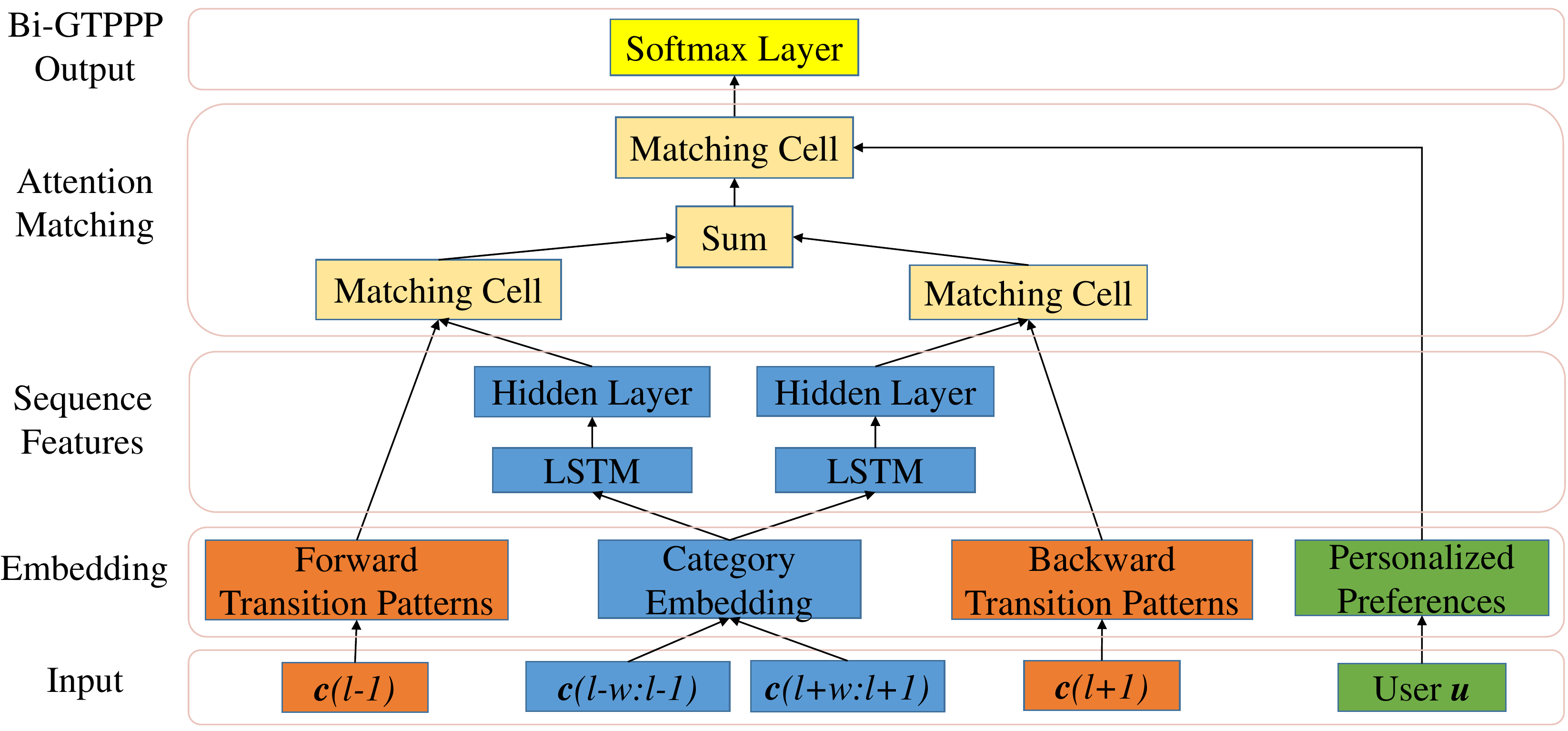}
		\caption{The proposed \model model}
		\label{fig:model}
	\end{center}
\end{figure*}
Different users have personal preferences which much affect the check-in behaviors of users. For example, most users usually watch a movie after dinner, but some users have a dinner after the movie. If we identify the missing category of the personalized users as the global transition pattern $dinner\rightarrow movie$, the model will have an erroneous identification. So here we need to capture users' personal preferences. If we use one-hot encoded user $\u\in\setR^N$ as input vectors, then the personal preference of user $u$ can be expressed as:
\begin{equation}
\p^u=f(\E_p^\top\u)\label{u},
\end{equation}
where $\E_p\in\setR^{N\times M}$, each row of which represents a user's personal preferences for $M$ categories. And the whole embedding matrix $\E_p$ can be seen as performing the latent factor modeling for all users' personal category preferences. We can initialize the matrix $\E_p$ via counting users' visiting history on all training data and fine-tune during model training.

Then, the same structure attention matching cell can be adopted to model how well the user's recent bi-directional check-in information and global transition patterns match his or her personal preferences:
\begin{equation}
\n^u_{l}=cell(\m^u_{l},\p^u).\label{personalized cell}
\end{equation}

Finally, the prediction of the $l$-th category the user $u$ has been can be computed as:
\begin{equation}
\o^u_l=softmax(\W_o\n^u_{l}),
\end{equation}
where $\W_o\in\setR^{M\times M}$ are the parameters in the softmax layer. The $\o^u_l$ is a distribution which indicates different probability of all possible candidate categories the user $u$ might visit $l$-th. And the categories corresponding to the k maximum probabilities are the top-k identifications for the missing category.

We present the final neural network architecture of \model in Figure \ref{fig:model}. First, bi-directional category sequence features are captured by two basic LSTM network, and then the inputs of categories and user are fed into embedding layer to capture bi-directional global transition patterns and personal preferences in Equations (\ref{gf}), (\ref{gb}) and (\ref{u}) respectively. Moreover, three same structure attention matching cells model how well the user's bi-directional category sequence features match the global transition patterns and how well the user's recent bi-directional check-in information and global transition patterns match his or her personal preferences in Equations (\ref{global cell}) and (\ref{personalized cell}) respectively. Finally, the softmax layer makes the identification for the missing $l$-th category.

We need to minimize the $cross\ entropy$ of predicted distribution and the actual distribution:
\begin{equation}
J(\theta)=-\frac{1}{S}\sum^S_{i=1}\sum^M_{j=1}y_{i,j}\log(o^u_{l,j}\vert \x_i,\theta),\label{equ:loss}
\end{equation}
where $S$ is the number of samples, $M$ is the number of categories, $\y_i\in\setR^M$ is the one-hot label of sample $\x_i$ and $\theta$ is the parameters set.

The training is performed in an end-to-end manner and the Equation (\ref{equ:loss}) is used to train globally the whole architecture.
Training is done through stochastic gradient descent over shuffled mini-batches with the
Adam \citep{kinga2015method} update rule.
\setcounter{footnote}{0}
\section{Experiments}
In this section, we perform experiments to evaluate the proposed \model model against various baseline methods on two real-world LBSNs datasets. 
We first introduce the datasets, baseline methods, implementation details and evaluation metrics of our experiments. Finally, we present our experimental results and analysis.

\subsection{Datasets}
The statistics of the two public LBSNs datasets are listed in Table \ref{tab:dataset}.
\begin{table}
  \centering
  \caption{Statistics of the two datasets.}
  \resizebox{0.9\linewidth}{!}{
    \begin{tabular}{cccccc}
    \toprule
    Dataset & \#user & \#POI & \#category & \#check\_in & \#Avg.check-in \\
    \midrule
    NYC & 1,083 &38333 & 251 & 227,428 & 210.0  \\
    TKY & 2,293 &61858 & 247 & 573,703 & 250.2  \\
    \bottomrule
    \end{tabular}%
    }
  \label{tab:dataset}
\end{table}%

\begin{itemize}
\item \textbf{NYC}\footnote{https://sites.google.com/site/yangdingqi/home/foursquare-dataset\label{nyc}}  \citep{yang2015modeling} is a dataset from Foursquare, which includes long-term (about 10 months) check-in data in New York city collected from April 2012 to February 2013.
\item \textbf{TKY}\textsuperscript{\ref{nyc}} \citep{yang2015modeling} is a dataset similar to NYC except from Tokyo.
\end{itemize}

We eliminate users with fewer than 10 check-ins in these two datasets. Then, we sort each user's check-in records according to timestamp order, taking the first $80\%$ as training set, the following $10\%$ for the validation set and the remaining $10\%$ for the test set. The users' history category information is intentionally removed as ground truth for testing the identification performance, this is consistent for all experiments for fair comparison.

\subsection{Baselines}
We compare the proposed method with counting based methods (Forward, Backward, TOP1, TOP2), traditional POI recommendation algorithms (PRME, PRME-G), neural network-based approaches (RNN, LSTM, GRU, STRNN, PACE, Bi-STDDP). Some earlier methods likes PMF \citep{Salakhutdinov2007Probabilistic}, FPMC \citep{rendle2010factorizing}, FPMC-LR \citep{cheng2013you} have been proved to be not as good as PRME-G \citep{feng2015personalized,liu2016predicting,he2017category}, so we don't compare them with our model. 

\begin{itemize}
\item \textbf{Forward}: The forward transition probability between categories is taken as the prediction for all users.
\item \textbf{Backward}: The backward transition probability between categories is taken as the prediction for all users.
\item \textbf{TOP1}: The most popular categories in the training set are selected as the prediction for all users.
\item \textbf{TOP2}: The most popular categories in the training set are selected as the prediction for each user.
\item \textbf{PRME}\citep{feng2015personalized}: User and POIs are embedded into the same latent space to capture the user transition patterns.
\item \textbf{PRME-G}\citep{feng2015personalized}: It takes the distance between destination POIs and recent visited ones into consideration on the basis of PRME.
\item \textbf{RNN}\citep{zhang2014sequential}: This is a neural network method which directly models the dependency on user's sequential behaviors into the click prediction process through the recurrent structure in RNN.
\item \textbf{LSTM}\citep{hochreiter1997long}: This is a special RNN model, which contains a memory cell and three multiplicative gates to learn long-term dependency.
\item \textbf{GRU}\citep{Cho2014Learning}: This is another special RNN model, which contains two gates and is simpler than LSTM.
 \item \textbf{STRNN}\citep{liu2016predicting}: This is a RNN-based model for next POI recommendation. It incorporates both the time-specific transition matrices and distance-specific transition matrices within recurrent architecture.
\item \textbf{PACE}\citep{yang2017bridging}: This is a deep neural
architecture that jointly learns the embeddings of users and POIs
to predict both user preference over POIs and various context associated with users and POIs.
\item \textbf{Bi-STDDP}\citep{xi2019modelling}: This is the state-of-the-art missing POI check-in identification model which captures bi-directional spatio-temporal dependence and users’ dynamic preferences.
\end{itemize}

\subsection{Implementation Details}
For all datasets we use: embedding dimension $d$ of 128, LSTM output space $h$ of 512, window width $w$ of 18, mini-batch size of 128 and learning rate of 0.001. All these values are chosen via a grid search on the NYC validation set.  
The parameters of $\E_p$ are initialized via counting users' visiting history on training data, 
and all other parameters in the neural network are initialized from glorot uniform distributions \citep{glorot2010understanding}. We do not perform any datasets-specific tuning except early stopping on validation sets.

\subsection{Evaluation Metrics}
To evaluate the performance of our proposed \model model and the baselines described above, we follow the existing works \citep{liu2016predicting} to use several standard metrics: \textbf{Recall@K}, \textbf{F1-score@K} and Mean Average Precision (\textbf{MAP}). 
Recall@K is 1 if the category visited appears in the top-K ranked list; otherwise is 0. The final Recall@K is the average value over all test ground truth instances. MAP is a global evaluation for ranking tasks, and it is usually used to evaluate the quality of the whole ranked lists. We report Recall@K and F1-score@K with K = 1, 5 and 10 in our experiments. The larger the value, the better the performance for all the evaluation metrics.
All the metrics reported in the experiment are the mean over five runs.

\begin{table*}[!th]
	\centering
	\caption{Evaluation of missing category identification in terms of Recall@K, F1-score@K and MAP. Bi-GTPPP-rand means that our model parameters $\E_p$ are initialized from glorot uniform distributions, and Bi-GTPPP-nonstatic means that parameters $\E_p$ are initialized via counting users' visiting history on training data and fine-tune during training. Underlined results indicate the best baselines over each dataset and metric. ``*'' indicates that the improvement is statistically significant compared with the best baselines at p-value $<$ 0.05 over independent samples t-tests.}
	\resizebox{1\linewidth}{!}{
	\begin{tabular}{@{}ccccccccc@{}}
		\toprule
		&     &Recall@1 & Recall@5 &Recall@10 & F1-score@1 & F1-score@5 & F1-score@10 & MAP \\
		\midrule
		\multirow{13}[8]{*}{NYC} & Forward & 0.1576  & 0.3593  & 0.4877  & 0.1576  & 0.1198  & 0.0887  & 0.2636  \\
		& Backward & 0.1566  & 0.3505  & 0.4868  & 0.1566  & 0.1168  & 0.0885  & 0.2601  \\
		& TOP1 & 0.0594  & 0.2888  & 0.4127  & 0.0594  & 0.0963  & 0.0750  & 0.1756  \\
		& TOP2 & 0.1688  & 0.3842  & 0.5017  & 0.1688  & 0.1281  & 0.0912  & 0.3009  \\
		\cmidrule{2-9}        & PRME & 0.1039  & 0.3249  & 0.4442  & 0.1039  & 0.1083  & 0.0808  & 0.2163  \\
		& PRME-G & 0.1180  & 0.3532  & 0.4818  & 0.1180  & 0.1177  & 0.0876  & 0.2334  \\
		\cmidrule{2-9}        & RNN & 0.1873  & 0.4954  & 0.6164  & 0.1873  & 0.1651  & 0.1121  & 0.3292  \\
		& LSTM & 0.2095  & 0.5191  & 0.6504  & 0.2095  & 0.1730  & 0.1183  & 0.3555  \\
		& GRU & 0.2162  & 0.5234  & 0.6493  & 0.2162  & 0.1745  & 0.1181  & 0.3596  \\
		& STRNN & 0.2331  & 0.5408  &  \underline{0.6689}  & 0.2331  & \underline{0.1803}  & \underline{0.1216}  & 0.3787  \\
		& PACE & 0.2330  & 0.5401  & 0.6675  & 0.2330  & 0.1800  & 0.1214  & 0.3769  \\
		& Bi-STDDP & \underline{0.2345}  & \underline{0.5409}  & 0.6677  & \underline{0.2345}  & \underline{0.1803}  & 0.1214  & \underline{0.3796}  \\
		\cmidrule{2-9}        & Bi-GTPPP-rand & 0.2425  & 0.5591  & 0.6702  & 0.2425  & 0.1864  & 0.1219  & 0.3898  \\
		& Bi-GTPPP-nonstatic & \textbf{0.2580*} & \textbf{0.5745*} & \textbf{0.6922*} & \textbf{0.2580*} & \textbf{0.1915*} & \textbf{0.1259*} & \textbf{0.4030*} \\
		\midrule
		\multirow{13}[8]{*}{TKY} & Forward & 0.3920  & 0.5950  & 0.6899  & 0.3920  & 0.1983  & 0.1254  & 0.4962  \\
		& Backward & 0.3924  & 0.5973  & 0.6919  & 0.3924  & 0.1991  & 0.1258  & 0.4965  \\
		& TOP1 & 0.3806  & 0.5376  & 0.6385  & 0.3806  & 0.1792  & 0.1161  & 0.4694  \\
		& TOP2 & 0.3967  & 0.6225  & 0.7013  & 0.3967  & 0.2075  & 0.1275  & 0.5109  \\
		\cmidrule{2-9}        & PRME & 0.3612  & 0.5265  & 0.6128  & 0.3612  & 0.1755  & 0.1114  & 0.4512  \\
		& PRME-G & 0.3638  & 0.5358  & 0.6204  & 0.3638  & 0.1786  & 0.1128  & 0.4558  \\
		\cmidrule{2-9}        & RNN & 0.4051  & 0.6773  & 0.7808  & 0.4051  & 0.2258  & 0.1420  & 0.5298  \\
		& LSTM & 0.4203  & 0.6956  & 0.7930  & 0.4203  & 0.2319  & 0.1442  & 0.5458  \\
		& GRU & 0.4189  & 0.6929  & 0.7912  & 0.4189  & 0.2310  & 0.1439  & 0.5445  \\
		& STRNN & \underline{0.4325}  & \underline{0.7071}  & \underline{0.8036}  & \underline{0.4325}  & \underline{0.2357}  & \underline{0.1461}  & \underline{0.5576}  \\
		& PACE & 0.4251  & 0.6930  & 0.8033  & 0.4251  & 0.2310  & \underline{0.1461}  & 0.5568  \\
		& Bi-STDDP & 0.4283  & 0.7065  & 0.8022  & 0.4283  & 0.2355  & 0.1459  & 0.5571  \\
		\cmidrule{2-9}        & Bi-GTPPP-rand & 0.4412  & 0.7088  & 0.8069  & 0.4412  & 0.2363  & 0.1467  & 0.5652  \\
		& Bi-GTPPP-nonstatic & \textbf{0.4454*} & \textbf{0.7211*} & \textbf{0.8146*} & \textbf{0.4454*} & \textbf{0.2404} & \textbf{0.1481} & \textbf{0.5721*} \\
		\bottomrule
	\end{tabular}
	}
	\label{tab:result}
\end{table*}%

\begin{figure*}[!ht]
	\begin{center}
			\includegraphics[width=\linewidth]{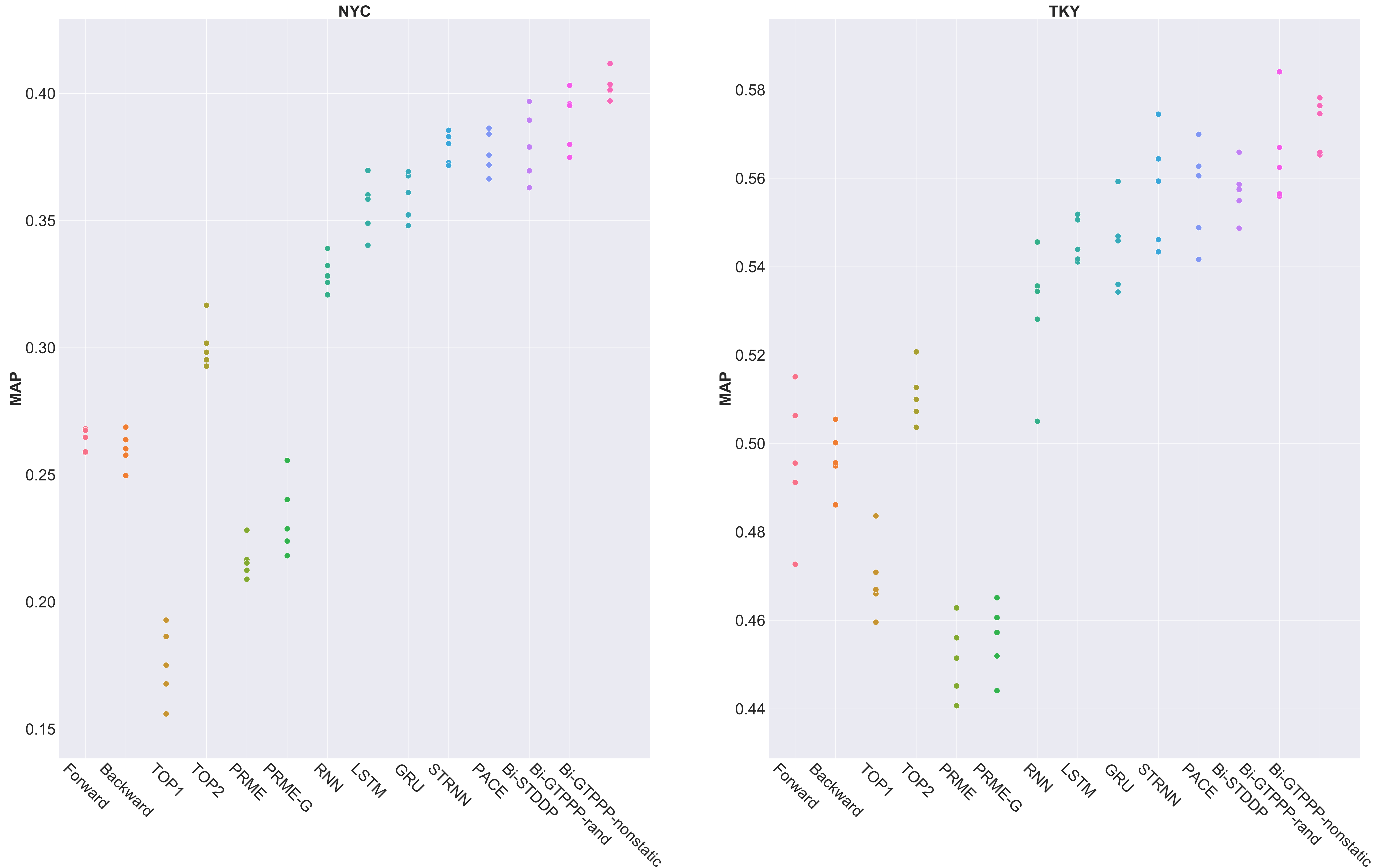} 
		\caption{The scatter plot of different methods in term of MAP over five runs.}
\label{fig:scatterplot} 
\end{center}
\end{figure*}

\begin{figure*}[!ht]
	\begin{center}
			\includegraphics[width=\linewidth]{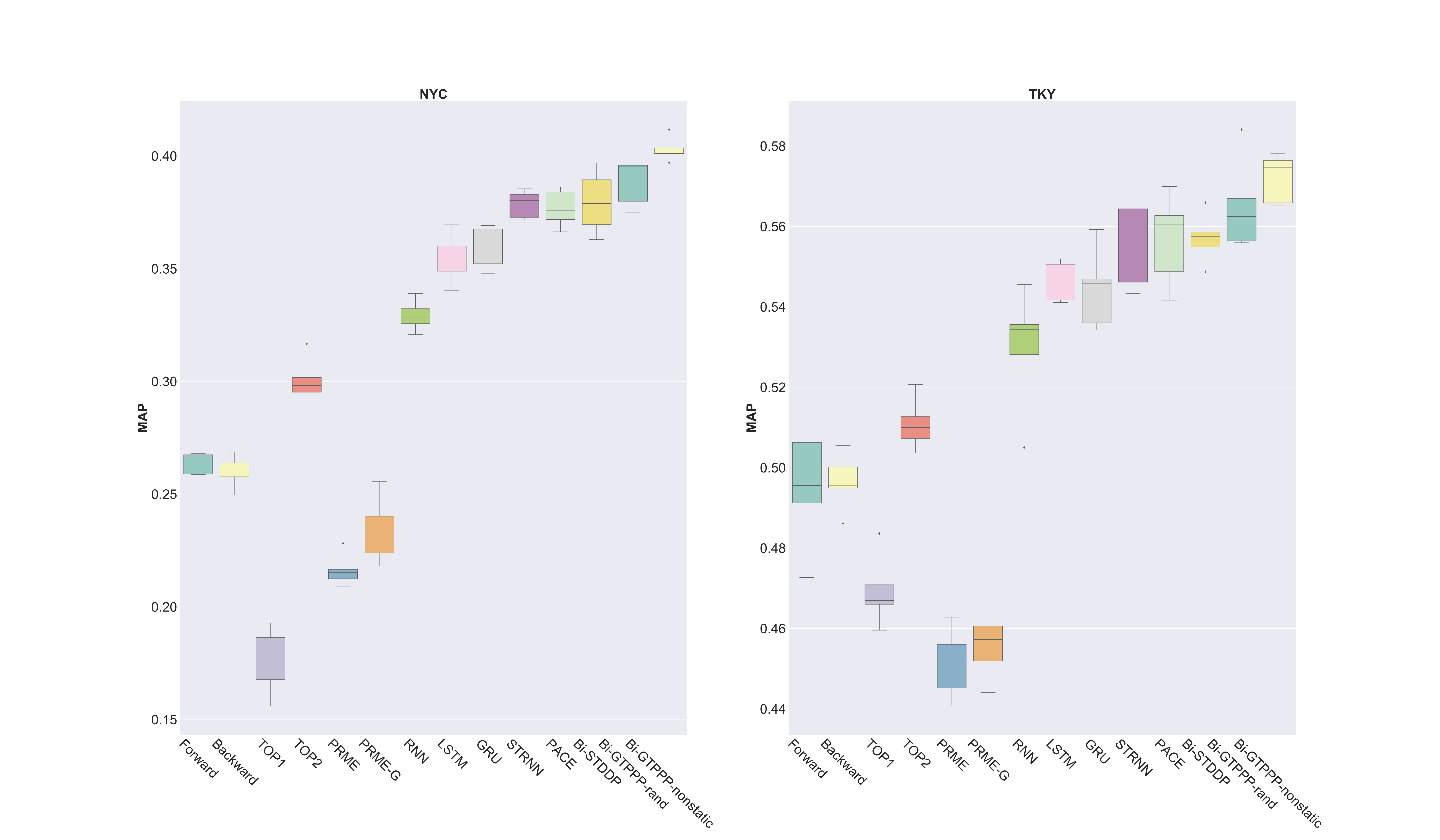} 
		\caption{The box plot of different methods in term of MAP over five runs.}
			\label{fig:boxplot} 
	\end{center}
\end{figure*}

\subsection{Performance Comparison and Discussion }
We present the experimental results evaluated by Recall@K and MAP on NYC and TKY datasets in Table \ref{tab:result}. 
Besides, we also present the scatter and box plots of different methods in term of MAP over five runs in Figure \ref{fig:scatterplot} and Figure \ref{fig:boxplot}.
From these results, we have the following insightful observations,
\begin{itemize}
	\item[-] The counting-based methods Forward and Backward have acceptable performances on all two datasets. Similarly, counting-based personalized TOP2 also has a good performance on NYC and TKY, and these results are even better than PRME and PRME-G. This makes sense that users' behavior patterns are usually regular and follow the long-tailed distribution. While the non-personalized TOP1 performs differently on two datasets, this seems to show that the users' preferences in Tokyo are more consistent than those in New York City.
	\item[-] PRME-G slightly improves the results comparing with PRME via incorporating distance information. And RNN-based methods (RNN, LSTM, GRU) obtain similar performance improvement over PRME-G because of their sequence modeling capability.
	\item[-] PACE predicts user preference over POIs, user context and POI context together to achieve further improvement over RNN-based methods.
	Another great improvement is achieved by STRNN. It incorporates both the time-specific transition matrices and distance-specific transition matrices within recurrent architecture in each layer. 
	Bi-STDDP achieves similar performance improvement over STRNN via incorporating both the bi-directional spatio-temporal dependence and users’ dynamic preferences, and they are the best methods among the baselines on the two datasets.
	However, the missing POI category check-ins identification is a little different from the missing POI check-in identification. POI check-ins are more dependent on spatio-temporal information (e.g., it is impossible that two POI check-ins of the same user are far apart in space distance, but the time interval is very short.), but POI category check-ins do not have the obvious spatio-temporal dependence due to POIs at different distances may belong to the same category. Therefore, the Bi-STDDP can not address the missing POI category identification well.
	\item[-]  Bi-GTPPP-rand outperforms the baseline methods over all evaluation metrics on all two datasets. And Bi-GTPPP-nonstatic further improves the performance via utilizing priori information and fine-tuning during training.
	On NYC dataset, the performance improvement of Bi-GTPPP-nonstatic on Recall@1, Recall@5, Recall@10 comparing with Bi-GTPPP-rand are $6.39\%$, $2.75\%$, $3.28\%$ respectively, and comparing with best baseline Bi-STDDP are $10.02\%$, $6.21\%$ and $3.67\%$ respectively which indicates that the Bi-GTPPP-nonstatic improves even more on the higher ranking list. Similar results can also be observed on TKY dataset.
	Besides, the baseline Bi-STDDP is designed for missing POI identification and captures bi-directional spatio-temporal dependence, which is unsuitable for the missing POI category identification task. Because POIs in different locations may belong to the same category, which results in the spatio-temporal dependence is poor in missing POI category identification task.
	As aforesaid, POI-oriented tasks usually suffer from huge search space due to the large amount of POIs, while the POI category can help filter candidate POIs and thus reduce the search space for efficiency. For the TKY dataset, the numbers of the POIs and categories are 61858 and 247, respectively, and one category corresponds to 250 POIs on average. If we adopt the top 10 categories, the recall@10 is 0.8146 and we can reduce the candidate POIs from 61858 to $10\times250=2500$, which is a 25 times reduction.
\end{itemize}
Overall, these improvements indicate the fact that the baseline methods fail to capture both global transition patterns and users' personal preferences, while our proposed \model can do this.

\subsection{Impact of Different Parts}
\begin{table}
	\centering
	\caption{Impact of forward and backward sequences on NYC dataset evaluated by Recall@K and MAP.}
	\resizebox{0.9\linewidth}{!}{
	\begin{tabular}{@{}ccccc@{}}
		\toprule
		&Recall@1 & Recall@5 & Recall@10 & MAP \\
		\midrule
		F-GTPPP & 0.2394  & 0.5584  & 0.6787  & 0.3866  \\
		B-GTPPP & 0.2403  & 0.5590  & 0.6785  & 0.3869  \\
		Bi-GTPPP & \textbf{0.2580} & \textbf{0.5745} & \textbf{0.6922} & \textbf{0.4030} \\
		\bottomrule
	\end{tabular}
	}
	\label{tab:direction}%
\end{table}%
\begin{figure*}[!ht]
	\begin{center}
		\subfigure[Fine-tuned Forward Transition Patterns]{ 
			\label{fig:pre}
			\includegraphics[width=0.31\linewidth]{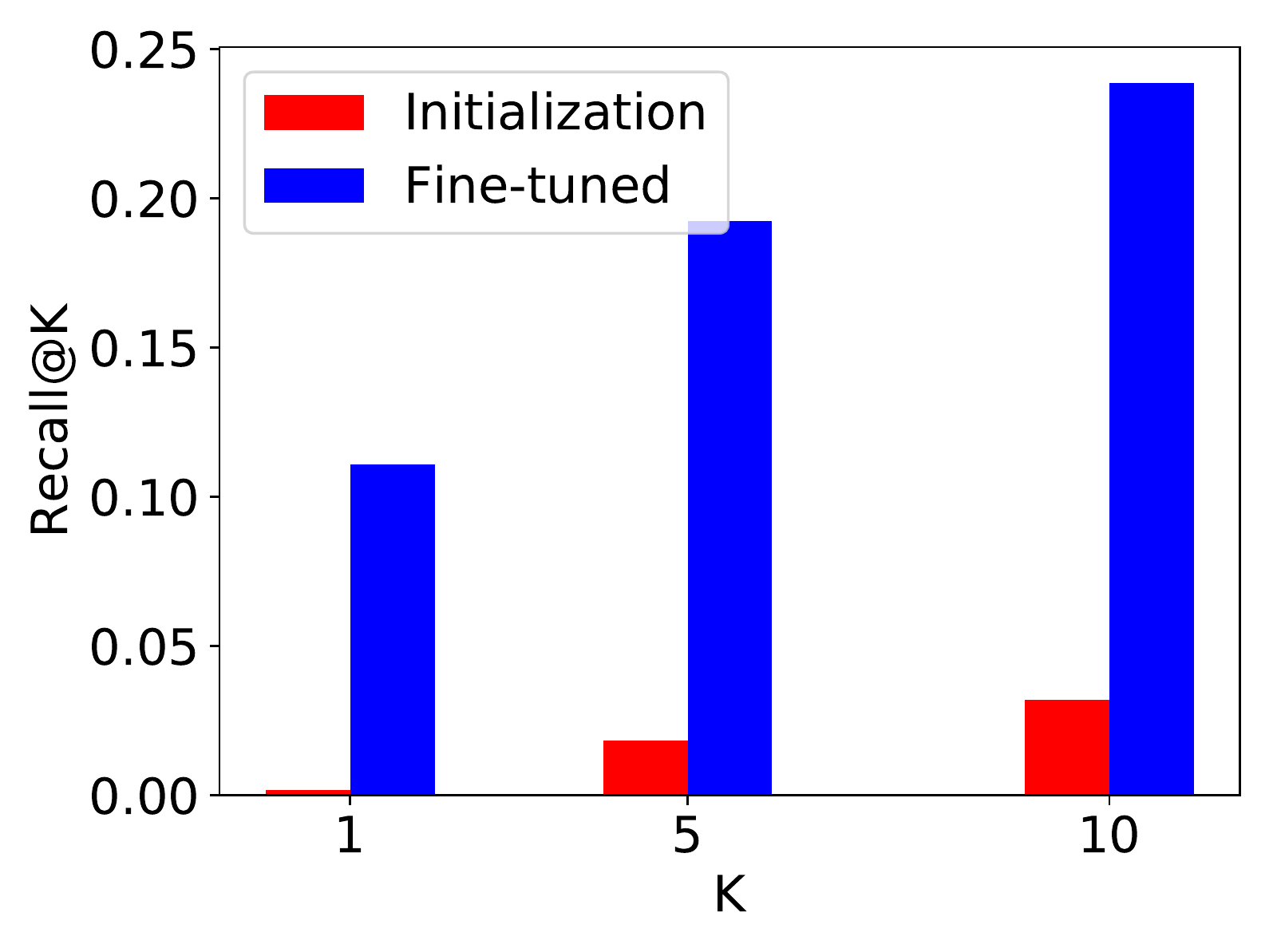}
		} 
		\subfigure[Fine-tuned Backward Transition Patterns]{ 
			\label{fig:post} 
			\includegraphics[width=0.31\linewidth]{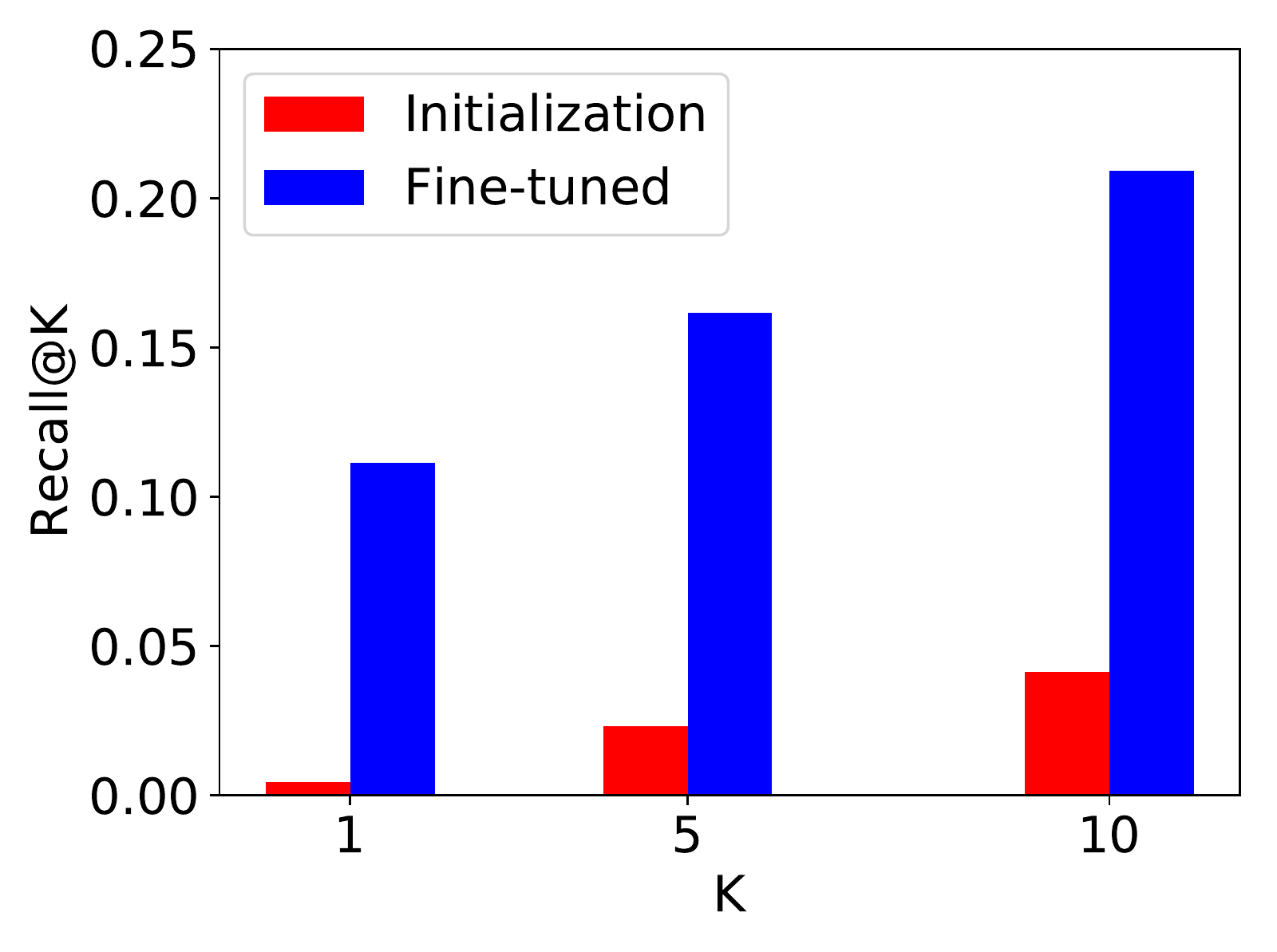}
		} 
		\subfigure[Fine-tuned Personal Preference]{ 
			\label{fig:user} 
			\includegraphics[width=0.31\linewidth]{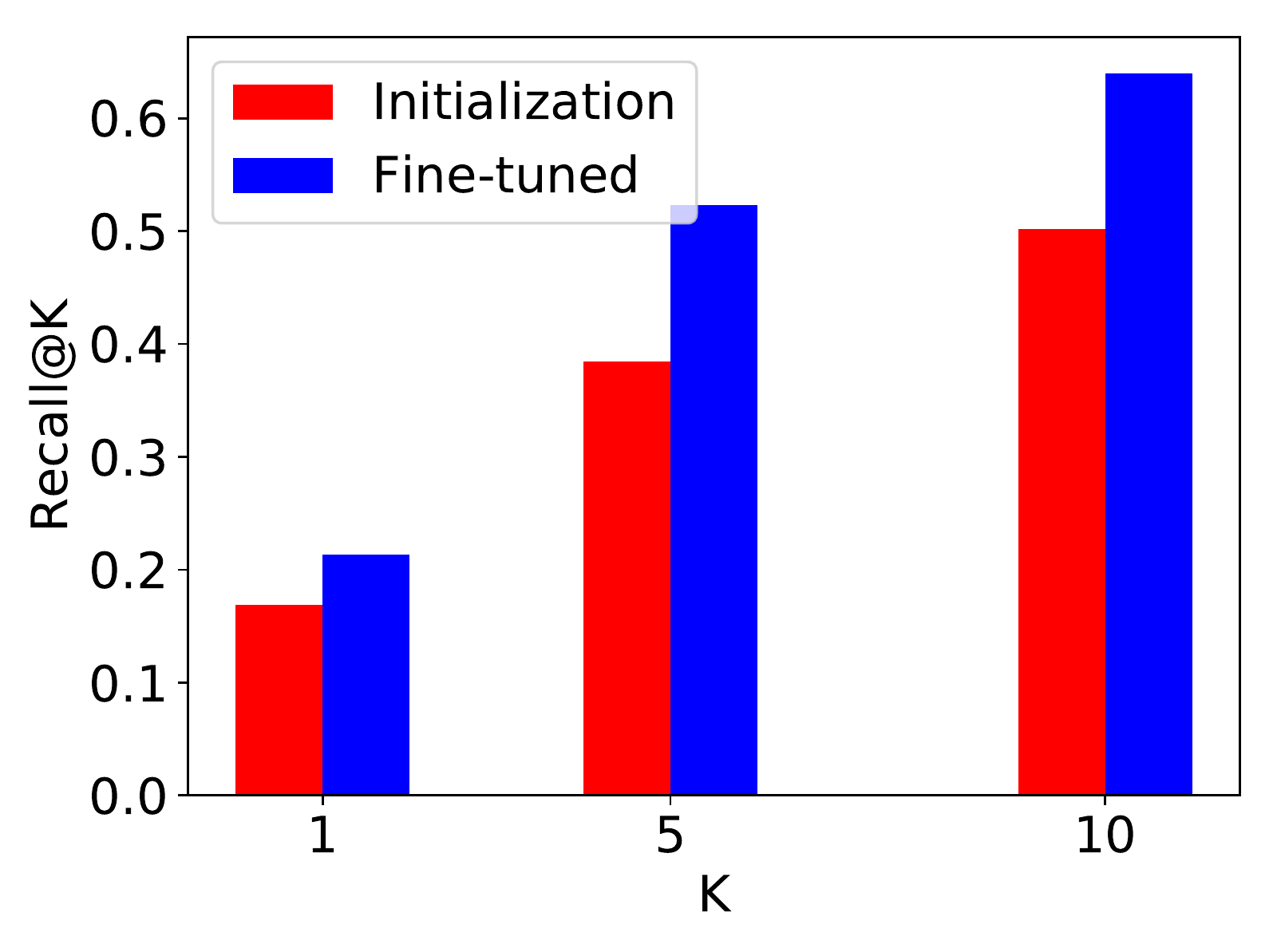}
		}
		\caption{Performance of fine-tuned bi-directional global transition patterns and personal preference on NYC dataset evaluated by Recall@K.}
		\label{fig:preference}
	\end{center}
\end{figure*}
\begin{figure*}[!ht]
	\begin{center}
		\subfigure[Impact of Embedding Dimension $d$]{ 
			\label{fig:embedding}
			\includegraphics[width=0.31\linewidth]{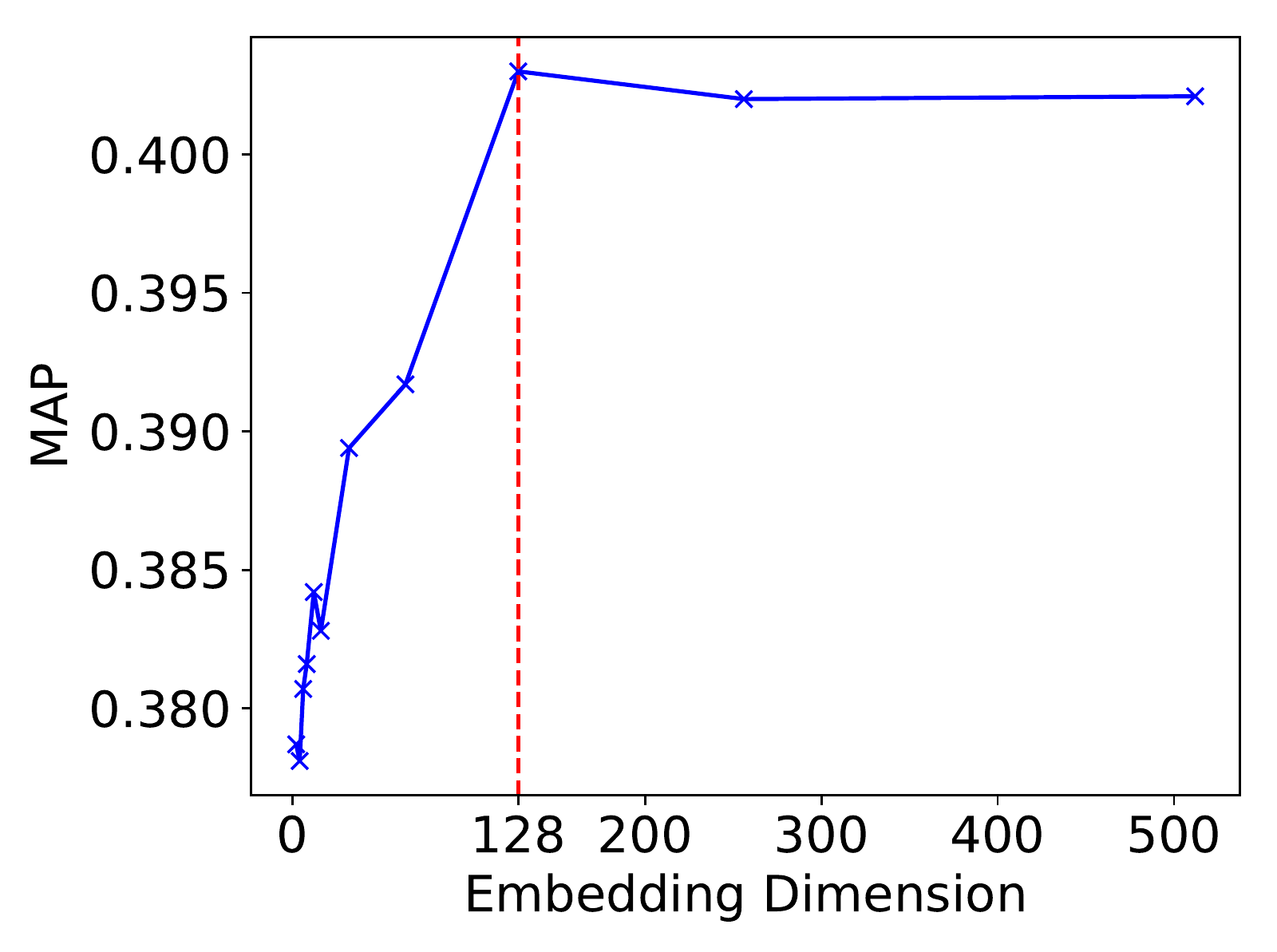} 
		} 
		\subfigure[Impact of LSTM Output Space $h$]{ 
			\label{fig:unit} 
			\includegraphics[width=0.31\linewidth]{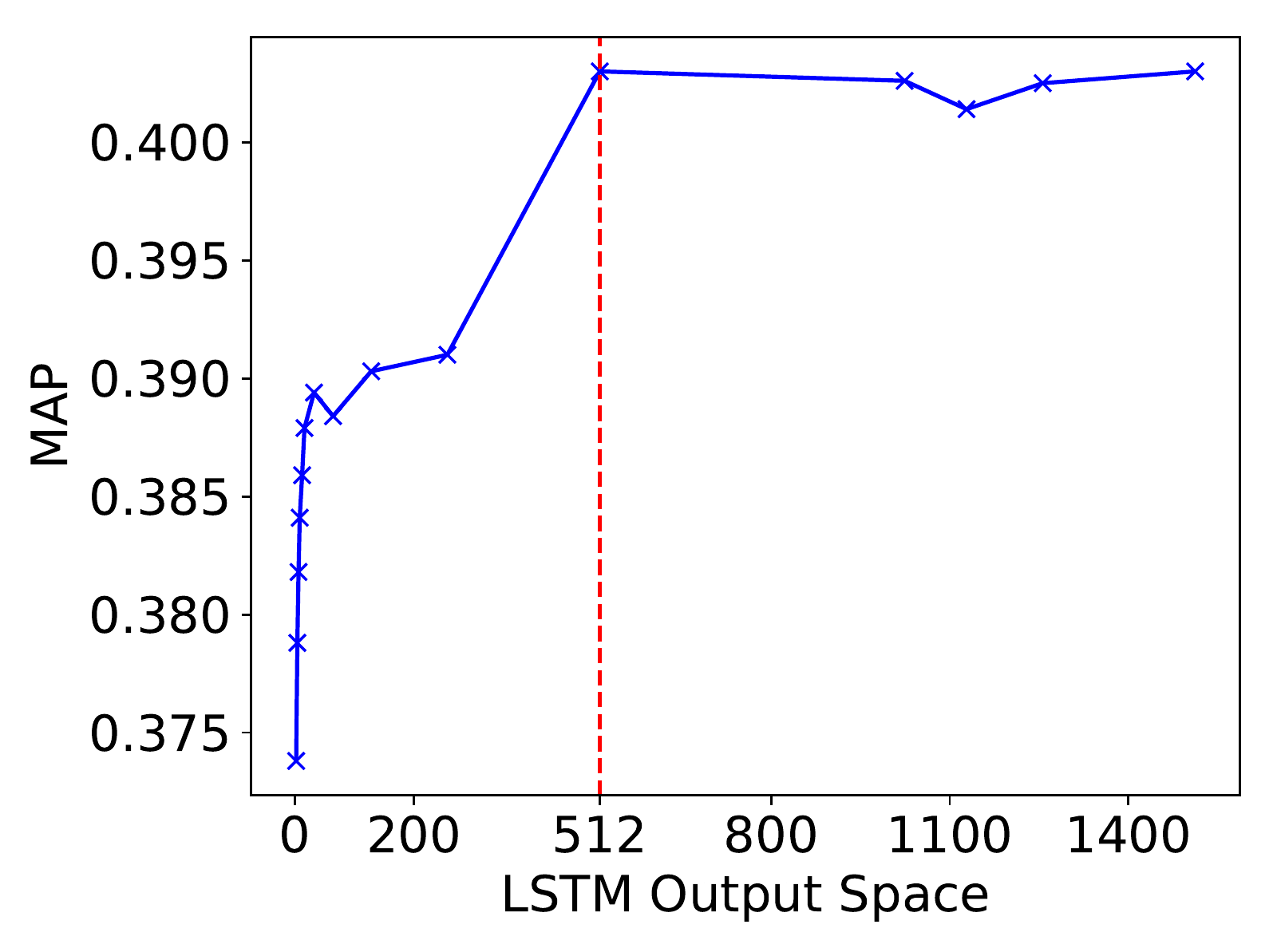} 
		} 
		\subfigure[Impact of Window Width $w$]{ 
			\label{fig:window} 
			\includegraphics[width=0.31\linewidth]{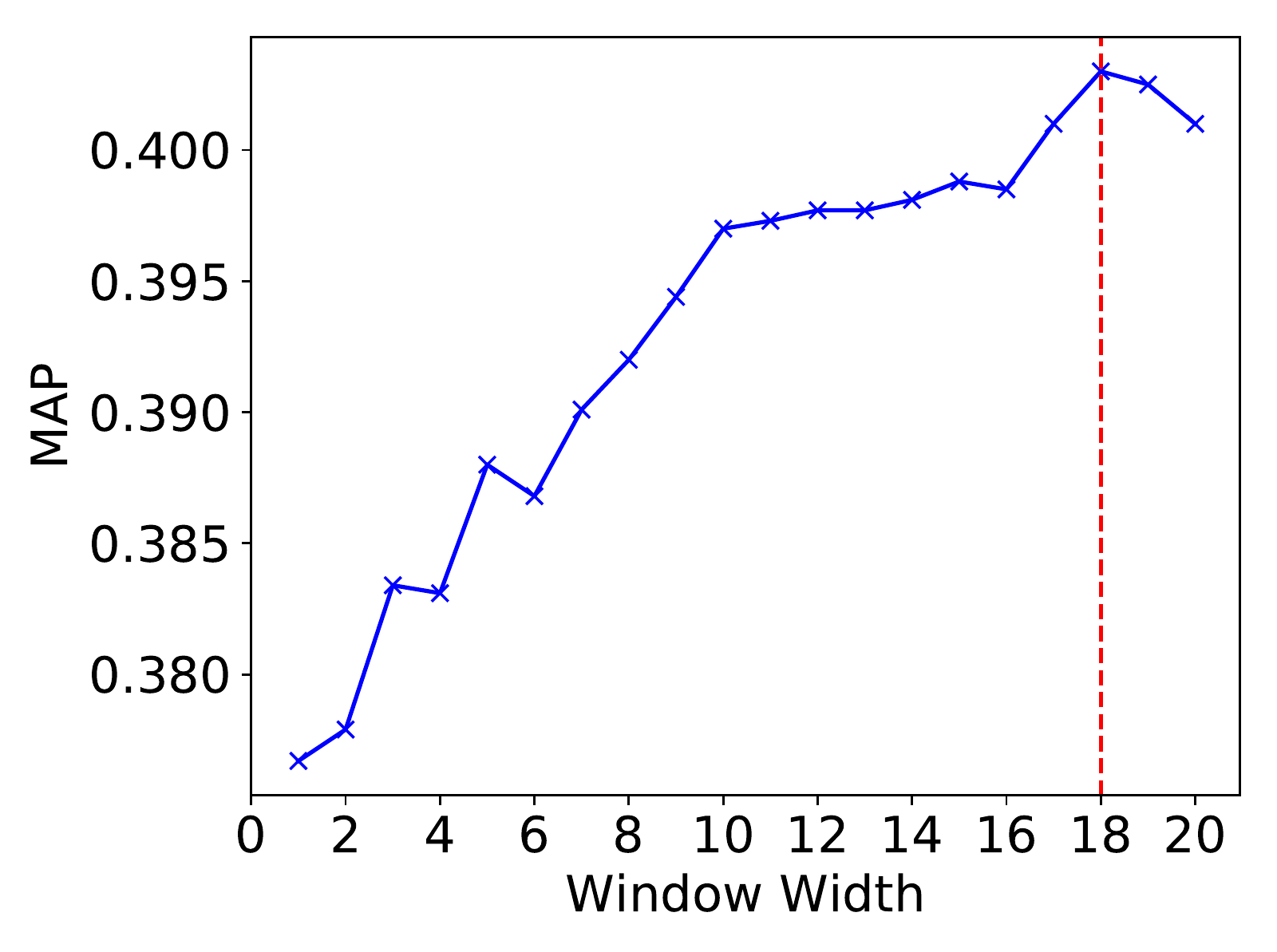} 
		} 
		\caption{Performance of \model with varying embedding dimension, LSTM output space and window width on NYC dataset evaluated by MAP.}
		\label{fig:parameter}
	\end{center}
\end{figure*}
In this subsection, we investigate the influence of the forward and backward sequences in our \model model. An intuitive feeling is that bi-directional sequences can bring more useful additional information and should have a better performance than single sequence. The results shown in Table \ref{tab:direction} confirm this. The \model achieves further improvement by utilizing  bi-directional sequences comparing with F-GTPPP and B-GTPPP which only utilize forward or backward information respectively. It must be noted that, although with the single sequence, our proposed model still achieves substantial performance improvement over various kinds of state-of-the-art methods. Based on this fact, we can say that our \model can be naturally extended to address next POI category recommendation and prediction tasks with competitive performance.

Next, we show that our \model model can capture bi-directional global transition patterns and user's personal preferences. The $\E_f$, $\E_b$ and $\E_p$ denote the forward transition embedding matrix, backward transition embedding matrix and user's personal preferences respectively which can be used directly for missing POI category identification. The Recall@K performance of initialization and fine-tuned $\E_f$, $\E_b$ and $\E_p$ is shown in Figure \ref{fig:preference}. We observe that fine-tuned $\E_f$ and $\E_b$ in Figure \ref{fig:pre} and \ref{fig:post} greatly improve the identification performance comparing with random initialized. And fine-tuned $\E_p$ in Figure \ref{fig:user} also performs better than initialization which counts users' visiting history. It turns out that bi-directional global transition patterns and user's personalized preferences can be captured in our novel model.

Besides, we can obviously see that $\E_p$ obtains more gain than $\E_f$ and $\E_b$. It is intuitive that user's personal preferences should play a more important role in the model than the bi-directional transition patterns. Because user's personal preference ($\E_p$) captures personal preferences for each user while the bi-directional transition patterns ($\E_f$ and $\E_b$) are global and non-personalized for all users, and personalized identification can obtain more performance improvement.

\subsection{Impact of Parameters}
Tuning model parameters is critical to the performance of the proposed model, such as the embedding dimension $d$, LSTM output space $h$ and window width $w$ in our \model model. Figure \ref{fig:embedding}, \ref{fig:unit} and \ref{fig:window} show the results under different settings of $d$, $h$ and $w$. We present the MAP performance of \model on NYC test set. Note that the best parameters are selected by grid search on NYC validation set, while the impact of parameters is evaluated on NYC test set. Validation and test results are similar under different settings.

The embedding dimension $d$ and LSTM output space $h$ have similar performance as shown in Figure \ref{fig:embedding} and \ref{fig:unit}. We observe that as the embedding dimension and LSTM output space increase, the performance of the model is quickly improved, and then becomes stable. 
The embedding dimension and LSTM output space are related to model complexity, smaller values are difficult to fit the data, and larger values result in more complex model and require more computing resources. Making a balance between performance and efficiency, $d=128$ and $h=512$ are proper parameters.

Figure \ref{fig:window} investigates the impact of window width on NYC dataset evaluated by MAP. From the experimental results, we observe that the performance first improves quickly with the increase of window width $w$ and then drops down gradually. The reason is that, smaller value of $w$ prunes too many useful history information, and larger value of $w$ brings little useful additional information and may even hurt the model performance by introducing noise information due to the long interval. We finally select $w=18$ as the window width.

According to Table \ref{tab:result} and Figure \ref{fig:parameter}, we can see that even without the best parameters, \model still outperforms other baseline methods. In a word, the performance of \model stays stable in a large range of values of parameters and is not very sensitive to embedding dimension, LSTM output space and window width.

\section{Conclusion and Outlook}
In this paper, we proposed a novel neural network model named \model to identify the missing category of POI where a user has visited  at any time in the past. The \model integrated bi-directional global transition patterns and personal preferences via delicately designed attention matching cells.
Specifically, the attention matching cell first bi-directionally models how well the check-in category information captured by LSTM network matches the global transition patterns.
And then another attention matching cell models how well the recent check-in category information and global transition patterns of users match their personal preferences.
Finally, the proposed model blue is evaluated on two large-scale real-world datasets and the results demonstrate the performance improvement of our proposed model compared with various kinds of state-of-the-art baseline methods.
Also, the proposed model can be naturally extended to address next POI category prediction task with competitive performance by only using forward sequence.

Several directions are available for future research in the area. First, the missing POI category identification task can be further explored and applied to a wider range of applications, such as the urban function zone dividing and urban planning, and even used for criminal activities analysis. Second, using POI category information to assist POI-oriented research will have a broader application scenario in various real-world applications. 
Third, the explanation and interpretability of POI-oriented research will become more and more important.
Finally, the identification performance can be improved by combining with other disciplines, for example, by learning relevant experience from nonlinear systems \citep{cite1,cite2,cite3}.

\section{Acknowledgements}
The research work supported by the National Key Research and Development Program of China under Grant No. 2018YFB1004300, the National Natural Science Foundation of China under Grant No. U1836206, U1811461, 61773361, the Project of Youth Innovation Promotion Association CAS under Grant No. 2017146.

\bibliographystyle{elsarticle-harv} 
\bibliography{nn}





\end{document}